\documentclass{article}

\usepackage{arxiv}

\usepackage[utf8]{inputenc} 
\usepackage[T1]{fontenc}    
\usepackage{hyperref}       
\usepackage{url}            
\usepackage{booktabs}       
\usepackage{amsfonts}       
\usepackage{nicefrac}       
\usepackage{microtype}      
\usepackage{lipsum}
\usepackage{algorithm}
\usepackage{algorithmic}
\usepackage[T1]{fontenc}
\usepackage{graphicx}
\usepackage{caption}
\usepackage{subcaption}

\usepackage{amsmath}
\usepackage{booktabs}

\title{Fair-by-design explainable models for prediction of recidivism}

\author{
  Eduardo Soares \\
 Lancaster University\\
  Lancaster, LA1 4WA, UK \\
 \texttt{e.almeidasoares@lancaster.ac.uk} \\
   \And
 Plamen Angelov \thanks{Honorary Professor, Technical University, Sofia, Bulgaria.} \\
 School of Computing and Communications\\
  LIRA Research Centre\\
  Lancaster University\\
  Lancaster, LA1 4WA, UK \\
  \texttt{p.angelov@lancaster.ac.uk} \\
}

\begin{document}
\maketitle

\begin{abstract}
Recidivism prediction provides decision makers with an assessment of the likelihood that a criminal defendant will reoffend that can be used in pre-trial decision-making. It can also be used for prediction of locations where crimes most occur, profiles that are more likely to commit violent crimes. While such instruments are gaining increasing popularity, their use is controversial as they may present potential discriminatory bias in the risk assessment. In this paper we propose a new fair-by-design approach to predict recidivism. It is prototype-based, learns locally and extracts empirically the data distribution. The results show that the proposed method is able to reduce the bias and provide human interpretable rules to assist specialists in the explanation of the given results.

\end{abstract}


\section{Introduction}

Predictive algorithms are becoming popular within the criminal justice system as they operate as risk assessment instruments. Such instruments may be used in pre-trial decision-making, prediction of locations where crimes most occur, profiles that are more likely to commit violent crimes, as well as to predict people who are likely to reoffend \cite{chouldechova2017fair,flores2016false}. In each of these cases, a high-risk misclassification may negatively impact on a criminal defendant’s outcome \cite{dressel2018accuracy,wadsworth2018achieving}. Therefore, it is highly important to guarantee that such instruments are free from discriminatory biases \cite{zeng2017interpretable}. Some state-of-the-art algorithms may provide unethical practices and inequitable outcomes for minorities \cite{johndrow2019algorithm}.

The Correctional Offender Management Profiling for Alternative Sanctions (COMPAS) risk assessment tool, has been developed in 1998, and since then has been used to assess more than 1 million offenders \cite{dressel2018accuracy}. In a study conducted by ProPublica \cite{dressel2018accuracy} that analyzed the efficacy of COMPAS on more than 7000 individuals arrested in Broward County, Florida between 2013 and 2014 it was found that the likelihood of a non-recidivating black defendant being assessed as high risk is nearly twice that of white defendants. African-american defendants who did not recidivate were incorrectly predicted to reoffend at a rate of 44.9\%. On the other hand, their white counterparts were incorrectly classified with 23.5\%. Moreover, white defendants who recidivated were incorrectly predicted as not high risk to reoffend with 47.7\%, African-americans who recidivated were incorrectly predicted as not high risk to reoffend with 28.0\%. These findings indicate that the COMPAS instrument has considerably higher false positive rates and lower false negative rates for black defendants than for white defendants \cite{dressel2018accuracy}.

In this paper we propose a method and algorithm that offers self-learning locally generative models that work together and require very light supervision in the form of few training labeled data samples. This is in sharp contrast to the traditional approach where learning is, in essence, only an averaging of the history. The proposed approach is based on using prototypes and learning locally around them extracting the empirical data distribution called typicality as well as the data density \cite{angelov2019empirical}. The proposed approach is recursive, non-iterative, non-parametric, thus computationally very lean. This adds to its efficiency in terms of time and computational resources. From the user perspective, the proposed approach is clearly understandable to human users since it can be represented in a linguistic $IF ... THEN$ form. It combines reasoning and logic with machine learning. It can also be presented as a deep neural network. Finally, it also has a statistical nature an empirical form of the pdf \cite{angelov2019empirical}.

In this paper we apply our proposed approach to the COMPASS data in order to provide more fair results, avoiding discrimination and bias.

The remainder of this paper is organized as follows: The methods and algorithms section introduces the proposed fair-by-design approach. The experimental data employed in the analysis and results are presented in the results section. Conclusion is presented in the last section of this paper. 

\section{Methods and Algorithms}

The prototype-based learning is the core of the proposed method which represents (their focal points are prototypes) locally valid generative models described by multimodal Cauchy distribution \cite{angelov2019empirical}. The meta-parameters are initialized with the first observed data sample. The proposed algorithm works per class. 

\begin{equation}
P \leftarrow 1;~~~\mu \leftarrow \Bar{x}_{i};
\end{equation}
where $\mu$ denotes the global mean of data samples of the given class. $P$ is the total number of the identified prototypes from the observed data samples.

Each class $C$ is initialized by the first data sample of that class:
\begin{equation}
\begin{split}
\mathrm{C}_{1} \leftarrow x_{1};~~~p_{1} \leftarrow \Bar{x}_{1};\\
S_{1} \leftarrow 1;~~~r_{1}\leftarrow r^*;
\end{split}
\end{equation}

where, $p_{1}$ is the prototype of $\mathrm{C}_{1}$; $S_{1}$ is the corresponding support (number of members); $r_{1}$ is the corresponding radius of the area of influence of $\mathrm{C}_{1}$.

The next step is to calculate the data density at $\Bar{x}_{i}$ and $p_{j}~(j=1,2,...,P)$ is a Cauchy function.

\begin{subequations}
\begin{eqnarray}
 D_{i}(\Bar{x}_{i})=\frac{1}{1+\frac{||x_{i}-p_{i}||^2}{1-||\mu_{i}||^2}}. \label{couchy}
\label{eqy}
\end{eqnarray}
\end{subequations}

Then the algorithm absorbs the new data samples one by one by assigning them to the nearest (in the feature space) prototype:

\begin{equation}
n^*=\operatorname*{argmin}_{j=1,2,...,P}(||\Bar{x}_{i}-p_{j}||^2)
\end{equation}

Because of this form of assignment, the shape of the data partitioning is of the so-called Voronoi tesselation type \cite{okabe2009spatial}. We call all data points associated with a prototype \textit{data clouds}, because their shape is not regular (e.g., hyper-spherical, hyper-ellipsoidal, etc.) and the prototype is not necessarily the statistical and geometric mean \cite{angelov2019empirical}. 

Then we check if the following condition \cite{angelov2019empirical} is met:
\begin{equation}
\begin{split}
\textit{\text{IF }}(D_{i}(\Bar{x}_{i})\geq\max_{j=1,2,...,P}D_{i}(p_j))~~
\textit{\text{OR }}~~(D_{i}(\Bar{x}_{i})\leq \min_{j=1,2,...,P}D_{i}(p_j))\\
\textit{\text{ THEN }} (add~a~new~data~cloud)\label{eq8}
\end{split}
\end{equation}

It means that $\Bar{x}_{i}$ is out of the influence area of $p_j$. Therefore, $\Bar{x}_{i}$ becomes a new prototype of a new \textit{data cloud} with meta-parameters initialized by equation (\ref{eqIN}).
\begin{equation}
\begin{split}
P \leftarrow P+1;~~~\mathrm{C}_{P} \leftarrow \{\Bar{x}_{i}\};
p_{P} \leftarrow \Bar{x}_{i};~~~S_{P} \leftarrow 1;
~r_{P}\leftarrow r_o;
\end{split}
\label{eqIN}
\end{equation}

A new data cloud is then formed around this new prototype. Otherwise, parameters of the nearest existing \textit{data cloud} are updated online. It has to be stressed that all calculations per data cloud are performed on the basis of data points associated with a certain data cloud only (i. e. locally, not globally, on the basis of all data points).
One of the strongest aspects of the proposed approach is its high level of interpretability which comes from its prototype-based, local generative models as well as as its ability to be expressed as a set of linguistic $IF ... THEN$ rule of the following type:

\begin{equation}
\begin{split}
\mathrm{R}:~~~\textit{\text{IF }}(x\sim p_1)~~\textit{\text{OR }}~~ ... ~~\textit{\text{OR }}~~( x \sim p_{P}) ~ ~
\textit{\text{ THEN }} (\textit{\text{Class } }c)
\end{split}
\end{equation}

\noindent where $\sim$ denotes association/similarity to the prototypes. 

The learning algorithm of the proposed method is summarized below.

~~
\hrule
\vspace{6pt}

\textbf{Learning Procedure}
\vspace{3pt}
\hrule
\vspace{4pt}
\begin{algorithmic}[1]
	\STATE While the new data sample of the the c$-th$ class $x_{c,k}$ is available
	\STATE \textbf{IF} k = 1
	\STATE ~~ $P_c \leftarrow 1$; $\mu \leftarrow x_{1}$; $C_{1} \leftarrow x_{1}$; $p_{1} \leftarrow x_{1}$; $S_{1} \leftarrow 1$; $r_{c,1}\leftarrow r_o$; 
	\STATE \textbf{ELSE}
	\STATE ~~Calculate $ D^f_{i}$ using equation (\ref{couchy});
	\STATE ~~Update ${p}_{j}~(j=1,2,...,P)$ using equation (\ref{eqIN});
	\STATE ~~~~ \textbf{IF} Condition (eq. \ref{eq8}) is satisfied \textbf{THEN}
	\STATE ~~~~~~ Add a new data cloud;
	\STATE ~~~~ \textbf{ELSE} 
	\STATE ~~~~~~Updated nearest data cloud;
	\STATE ~~~~ \textbf{END}
	\STATE \textbf{END}
	
\end{algorithmic}
\vspace{4pt}
\hrule
\vspace{6pt}

\section{Results}

The analysis and results provided in this paper are based on the Broward County data made publicly available by ProPublica \cite{dressel2018accuracy}. This data set contains COMPAS recidivism risk decile scores, 2-year recidivism outcomes, and a number of demographic and crime-related variables on individuals who were scored in 2013 and 2014.

We compare the overall accuracy and groups accuracy with results provided by \cite{dressel2018accuracy}. We also compare results on false positives (a defendant is predicted to recidivate but they do not) and false negatives (a defendant is predicted to not recidivate but they do). It is important to highlight that the proposed approach uses weak supervision (only, 10\% of the available data was used to train the model).

As shown in Table \ref{results}, the proposed approach can obtain a better performance in terms of overall accuracy than its competitors. Moreover, it is important to highlight that the proposed approach works per class, in parallel, therefore, it can obtain a more balanced result than other state-of-the-art approaches. As shown in Table \ref{results}, the proposed approach was able to reduce the false positive rate for black people from $31.6\%$ (best case with NL-SVM) to $30.2\%$, additionally, the false negative rate, when a defendant is predicted to not recidivate but they do, for white population has decreased  from $46.1\%$ in the best scenario with $LR_2$ to $29.6\%$ in the proposed approach. 

\begin{table}[H]
  \caption{Experimental results}
  \label{sample-table}
  \centering
  \begin{tabular}{llllll}
    \toprule
    Results    & $LR_7$\cite{dressel2018accuracy}     & $LR_2$\cite{dressel2018accuracy} & NL-SVM \cite{dressel2018accuracy}& COMPAS \cite{dressel2018accuracy} & Proposed Approach\\
    \midrule
    Accuracy (overall) & 66.6\%  & 66.8\%  & 65.2\%  & 65.4\% & \textbf{67.7\%}\\
    Accuracy (black)    & 66.7\%& 66.7\%& 64.3\% &    63.8\% & \textbf{67.90\%}  \\
    Accuracy (white)  & 66.0\%     & 66.4\% & 65.3\% & 67.0\% & \textbf{68.5\%} \\
    False positive (black) & 42.9\%   & 45.6\% & 31.6\% &  44.8\%& \textbf{30.2\%} \\
    False positive (white)   & 25.3\% & 25.3\%  &  20.5\% & 23.5\%& 36.8\%\\
    False negative (black)   & 24.2\%       & 21.6\% & 39.6\% & 28.0\%& 34.1\%  \\
    False negative (white)   & 47.3\%      & 46.1\%  & 56.6\%  & 47.7\% & \textbf{29.6\%}\\
    \bottomrule
    \label{results}
  \end{tabular}
\end{table}

\begin{figure}[h]
	\begin{center}
		{\includegraphics[scale=0.32]{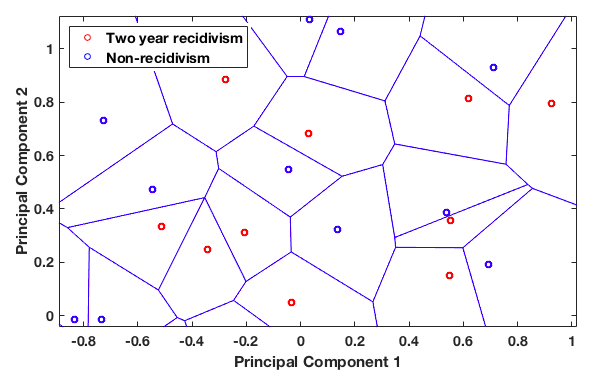}
		\caption{Voronoi Tesselation for the identified prototypes} \label{fair}

}	\end{center}
\end{figure}

The proposed approach is prototype-based and it learns locally around the prototypes extracting the empirical data distribution called typicality \cite{angelov2019empirical} as well as the data density. Figure \ref{fair} shows the prototypes identified during the learning phase of the proposed method.

The rule-based characteristic of the proposed method allows interpretability and explainability of the data. The prototypes identified for the `Two year recidivism for black people' rule are demonstrated on Table \ref{TableProt}. 
	\begin{table}[h]

	\small \caption{Identified Prototypes for the `Two year recidivism for black people' rule}
	\begin{center}
		\begin{tabular}{c|ccccccccc}
			\hline
		    Features &$p_1$ &$p_2$& $p_3$& $p_4$&$p_5$&$p_6$&$p_7$&$p_8$&$p_9$\\
			\hline
			Number of Priors (NP) & 14& 13 & 8 & 8 & 10& 7& 2& 0 &21   \\
			Score Factor (SF)& 1& 1 & 0 & 1 & 0& 1& 0& 0&1  \\
			Age above 45 (A45)& 0& 1 &0 & 0& 0& 0& 0& 0&0   \\
			Age below 25 (A25)& 0& 0 & 0 & 0& 0& 0& 0& 1&0  \\
			African American (AA)& 1& 1 & 1 &1 & 0& 1& 1& 1&1  \\
			Female (F)& 0& 0 & 0& 0& 1& 1& 1& 0&0  \\
			Misdemeanor (M)& 0& 0 & 0 &1 & 0& 0& 1& 1&0  \\
			\hline
		\end{tabular}
		\label{TableProt}
	\end{center}
\end{table}

The rule generated is as follows: 
\begin{equation}
\begin{split}
R: ~ IF ~ (\begin{bmatrix}
NP 
\\SF 
\\ A45 
\\ A25
\\  AA
\\ F
\\ M
\end{bmatrix} \sim \begin{bmatrix}
 14
\\ 1
\\  0
\\  0
\\  1
\\  0
\\ 0
\end{bmatrix}  ) ~OR~ (\begin{bmatrix}
NP 
\\SF 
\\ A45 
\\ A25
\\  AA
\\ F
\\ M
\end{bmatrix} \sim \begin{bmatrix}
 13
\\ 1
\\   1
\\ 0
\\  1
\\ 0
\\  0
\end{bmatrix}  ) ~OR...OR~ (\begin{bmatrix}
NP 
\\SF 
\\ A45 
\\ A25
\\  AA
\\ F
\\ M
\end{bmatrix} \sim \begin{bmatrix}
 21
\\ 1
\\   0
\\  0
\\   1
\\ 0
\\  0
\end{bmatrix}  ) 
~~THEN ~\text{`Two year recidivism' }
		\label{FigFinalRule}
\end{split}	
\end{equation}

These characteristics favour fairer results than traditional approaches that use only an averaging of the history of the data, and may ignore relevant information about individuals. In contrast, the proposed approach works locally building multiple models that have higher chance to capture more diverse data distribution.

\section{Conclusion}

In this paper we propose a new fair-by-design method to predict recidivism. It learns locally with very light supervision (using 10\% of labeled data) and extracts the empirical data distribution from the data. The results demonstrated that the proposed method is able to produce fairer and more accurate results than the traditional approaches. Moreover, human interpretable rules are provided to assist specialists in the explanation of the generated results.  

\bibliographystyle{unsrt}  
\bibliography{references}  


\end{document}